\documentclass[conference]{IEEEtran}
\IEEEoverridecommandlockouts
\usepackage{cite}
\usepackage{amsmath,amssymb,amsfonts}
\usepackage{graphicx}
\usepackage{textcomp}
\usepackage{xcolor}
\usepackage{comment}
\usepackage{mathrsfs}
\usepackage{setspace}
\usepackage{amssymb}
\usepackage{amsmath}
\usepackage{amsthm}
\usepackage{enumitem}
\usepackage{subcaption}
\usepackage{color}
\newtheorem{thm}{Theorem}
\newtheorem{mydef}{Definition}

\usepackage[linesnumbered,ruled,vlined]{algorithm2e}
\usepackage{algpseudocode}

\renewcommand{\baselinestretch}{1} 

\def\BibTeX{{\rm B\kern-.05em{\sc i\kern-.025em b}\kern-.08em
    T\kern-.1667em\lower.7ex\hbox{E}\kern-.125emX}}

\begin{document}

\title{Data-Free Evaluation of User Contributions in Federated Learning
}

\author{\IEEEauthorblockN{Hongtao Lv\IEEEauthorrefmark{1}, Zhenzhe Zheng\IEEEauthorrefmark{1}, Tie Luo\IEEEauthorrefmark{2}, Fan Wu\IEEEauthorrefmark{1}, Shaojie Tang\IEEEauthorrefmark{3}, Lifeng Hua\IEEEauthorrefmark{4}, Rongfei Jia\IEEEauthorrefmark{4}, Chengfei Lv\IEEEauthorrefmark{4}}
\IEEEauthorblockA{\IEEEauthorrefmark{1}
Shanghai Jiao Tong University
\IEEEauthorrefmark{2}
Missouri University of Science and Technology
\\
\IEEEauthorrefmark{3}
University of Texas at Dallas
\IEEEauthorrefmark{4}
Alibaba Group\\
Email: \{lvhongtao, zhengzhenzhe\}@sjtu.edu.cn, tluo@mst.edu, wu-fan@sjtu.edu.cn, \\ shaojie.tang@utdallas.edu, \{issac.hlf, rongfei.jrf, chengfei.lcf\}@alibaba-inc.com
}
\thanks{This work was supported in part by China NSF grant No. 62025204, 62072303, 61972252, 61902248, and 61972254, in part by the National Science Foundation (NSF) under Grant CNS-2008878, in part by Shanghai Science and Technology fund 20PJ1407900, in part by Alibaba Group through Alibaba Innovation Research Program, and in part by Tencent Rhino Bird Key Research Project. The opinions, findings, conclusions, and recommendations expressed in this paper are those of the authors and do not necessarily reflect the views of the funding agencies or the government.}

\thanks{Z. Zheng is the corresponding author.}
}

\maketitle

\begin{abstract}
Federated learning (FL) trains a machine learning model on mobile devices in a distributed manner using each device's private data and computing resources. A critical issues is to {\em evaluate individual users' contributions} so that (1) users' effort in model training can be compensated with proper incentives and (2) malicious and low-quality users can be detected and removed.
The state-of-the-art solutions require a representative test dataset for the evaluation purpose, but such a dataset is often unavailable and hard to synthesize. In this paper, we propose a method called {\em Pairwise Correlated Agreement} (PCA) based on the idea of {\em peer prediction} to evaluate user contribution in FL without a test dataset. PCA achieves this using the statistical correlation of the model parameters uploaded by users.
We then apply PCA to designing (1) a new federated learning algorithm called Fed-PCA, and (2) a new incentive mechanism that guarantees truthfulness. 
We evaluate the performance of PCA and Fed-PCA using the MNIST dataset and a large industrial product recommendation dataset. The results demonstrate that our Fed-PCA outperforms the canonical FedAvg algorithm and other baseline methods in accuracy, and at the same time, PCA effectively incentivizes users to behave truthfully. 

\end{abstract}

\begin{IEEEkeywords}
Peer prediction, correlated agreement.
\end{IEEEkeywords}

\section{Introduction}
\label{intro}
Recent years have seen a large variety of mobile applications that have changed or improved our ways of communication, shopping, commuting, traveling, and lifestyle. 
To provide personalized services in these applications, machine learning techniques have been increasingly adopted. For that purpose, a common practice is to upload user data to a central server or cloud, which then trains a machine learning model to make predictions such as product recommendation. This centralized approach evokes many privacy concerns as users' sensitive data could be eavesdropped by malicious parties during data transmission, or be misused by an untrusted server. To address this privacy issue, a new learning paradigm called \emph{Federated Learning} (FL) \cite{mcmahan2017aistat} was proposed to perform distributed machine learning over a large number of devices without requiring data to leave the devices or data owners.
In the training process, each user trains a local model using her own data on her own device, and uploads the local model parameters instead of the original data to the server.
The server then aggregates the received models into a global model and distributes the global model back to the users. The above steps repeat until the global model converges. 
By doing so, federated learning preserves user privacy and reduces network traffic. As a result, it has attracted substantial attention from both academia and industry recently.

Unlike traditional distributed machine learning~\cite{qiu2016survey}, in which the machines for model training are fully controlled by a central server, federated learning works with autonomous mobile users who decide by themselves whether and how to participate in model training. This makes FL vulnerable to selfish and malicious users who may manipulate the training process such as falsifying the model parameters or send random models without any training effort. Therefore, a critical issue in FL is to \emph{evaluate individual users' contributions} in the model training process so that strategic users can be detected and truthful users can receive rewards proportional to their real contributions.

A popular approach, as recently proposed in~\cite{liu2020fedcoin, jia2019towards, wang2019measure}, uses the {\em Shapley value} to measure user contribution in federated learning.  
This approach computes the marginal increase of average accuracy of the model due to the addition of data points contributed by a user in model training. 
However, this has two major drawbacks: 1) computing the Shapley value involves permutation operation which has an exponential computational complexity; even though some approximate methods have been proposed, the computation is still costly~\cite{jia2019towards}; 2) more importantly, it relies on a representative test dataset to evaluate the model accuracy, but such a dataset is rarely available because no one knows which dataset perfectly mimics the distribution of future unseen data. 

To overcome these issues, we propose a {\em data-free} approach to evaluate user contribution in FL. This approach only uses users' uploaded models and does not need any extra training or test data. However, there are several key challenges. First, the uploaded models (typically neural networks) have complex structures and the correlation among model parameters is too intricate to express. Second, there may exist dishonest or malicious users who falsify their data or even directly manipulate model parameters \cite{lin2019free}, and detecting such behavior is hard because the server does not have access to user data.
These challenges set the problem of evaluating user contribution in FL distinct from the data quality evaluation problem in mobile crowdsensing \cite{iotj19crossval} and crowdsourcing~\cite{gong2018incentivizing}.

To this end, we propose a {\em Pairwise Correlated Agreement} (PCA) method to evaluate user contribution in FL. The basic idea is that, although the uploaded models are different among mobile users due to their non-i.i.d. (non-independent and identically distributed) data~\cite{zhao2018federated}, we can still extract certain internal correlation between the models {\em pairwise}, and exploit the correlation using an idea based on \emph{peer prediction} \cite{shnayder2016informed}.
More specifically, we characterize user contribution by how much a model uploaded by a user can predict the models uploaded by the others via the internal correlations.

With our PCA method, we apply it to two fundamental aspects of FL. First, we design a new {\em model aggregation} algorithm called Fed-PCA, which uses the user contribution evaluation results as the model weights for model aggregation performed by the server. This Fed-PCA algorithm accounts for {\em quality} of data, rather than just the user-reported data size (quantity); as a result, it offers significant potential to improve the accuracy of the aggregated model, and {\em counter the falsification of user-reported data size}.
Second, we apply PCA to designing an incentive mechanism which is strategy-proof to the following undesirable user behaviors: (i) {\em free riding}, where a user randomly generates model parameters without performing the actual model training, and (ii) {\em overly privacy-preserving}: adding excessive noises to model parameters and thus substantially degrading model prediction performance.

In summary, our main contributions are as follows:
\begin{itemize}[leftmargin=*]
\item We propose a Pairwise Correlated Agreement (PCA) method to evaluate user contributions in federated model training without using a test dataset. For motivation purposes, we also demonstrate the importance of contribution evaluation by showing the potential harm of strategic user behaviors on federated model training using our experiments conducted on an industrial dataset. 

\item Based on PCA, we design an algorithm called Fed-PCA which uses the user contributions computed by PCA as the model weights in model aggregation. 
We also apply PCA to designing a strategy-proof incentive mechanism which resists two types of common strategic behaviors of users: free riding and overly privacy-preserving.

\item We conduct extensive experiments with the widely used MNIST dataset and an industrial dataset obtained from Taobao, one of the largest commercial mobile recommender system in China. 
The results clearly demonstrate the effectiveness of PCA in detecting users' strategic behaviors. In addition, counter-intuitively, our Fed-PCA achieves equal or even better prediction accuracy as compared to FedAvg on different datasets, without knowing the training data size of each user.
\end{itemize}

\section{Preliminaries}\label{prel}

\subsection{Federated Learning}
In the canonical FL framework \cite{mcmahan2017aistat}, there is a set of users (clients) $U$ and a central server. Each user $i$ has a private local dataset $D^i$, such as the historical dataset of click behaviors related to goods or videos in mobile recommender systems. In each round $t$, $k$ users are randomly chosen to participate in the model training, and are denoted by the set $\{1,...,k\}$. At the beginning of round $t$, the server sends the global model $\boldsymbol {M}_{t-1}$ to the selected $k$ users, and each user trains a new local model $\boldsymbol {M}_t^i$ using her own data $D^i$, and uploads the model update $\boldsymbol {x}^i_t = \boldsymbol {M}^i_{t}-\boldsymbol {M}_{t-1}$ to the server. Besides the model update, each user also reports $n_t^i$, the size of training data used at round $t$. The server then calculates the weight of each user $i$ as $w_t^i = n_t^i / \sum^k_{j=1} n_t^j$, and aggregates the model updates by $\overline{\boldsymbol {x}}_t = \sum^k_{i=1} w_t^i \boldsymbol {x}^i_{t}$. Finally, the global model is updated as $\boldsymbol {M}_t = \boldsymbol {M}_{t-1} + \overline{\boldsymbol {x}}_t$, which is sent back to users. The above process repeats until a stopping condition (e.g. a certain number of rounds) is met. As model compression with the quantization technique (e.g., \cite{agarwal2018cpsgd}) is widely used in FL to reduce communication cost, we take each parameter of the model, indexed by $p$, to be discrete and finite without loss of generality, i.e., $x^i_{t,p}\in\{1,2,...,h\}$. Note, however, that our proposed approach can be used for continuous parameter values as well, in which case we can perform an additional quantization of the model update on the server only for the contribution evaluation phase, and it does not affect the model aggregation (i.e., the model update of each user $i$ does not need to be quantized in the model aggregation).

\subsection{Undesirable User Strategies}
\label{strategy}
In an ideal FL framework, each user would  participate in the model training, upload her model truthfully. However, there may exist strategic users who can manipulate this process to achieve their own interests, e.g., by uploading a fake model update $\hat{\boldsymbol {x}}^i\neq \boldsymbol {x}^i$ where $\boldsymbol {x}^i$ is the true model update (omitting subscript $t$ for notation simplicity). We account for two main types of strategic user behaviors in FL: 1) free riding  \cite{lin2019free},  which generates random model parameters without actually training to save training costs, such as computing power and storage; and 2) overly privacy-preserving, which adds excessive noises to the model parameters for  privacy protection~\cite{abadi2016deep,agarwal2018cpsgd,bu2019deep}). 

\begin{figure}[t]
\centering
\subcaptionbox{}{\includegraphics[width=0.24\textwidth,trim=0 0 160 60,clip]{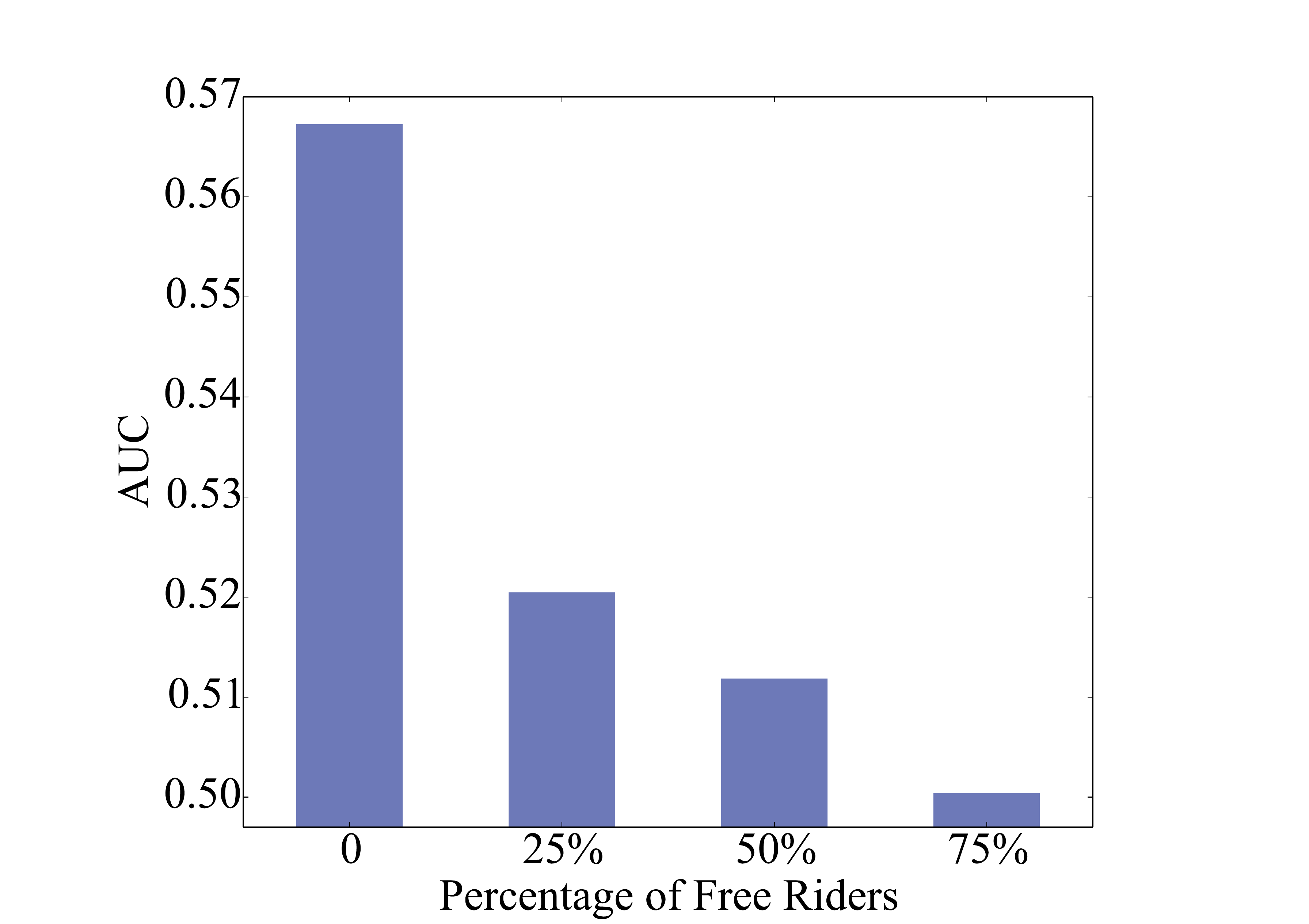}}%
\hfill
\subcaptionbox{}{\includegraphics[width=0.24\textwidth,trim=0 0 160 60,clip]{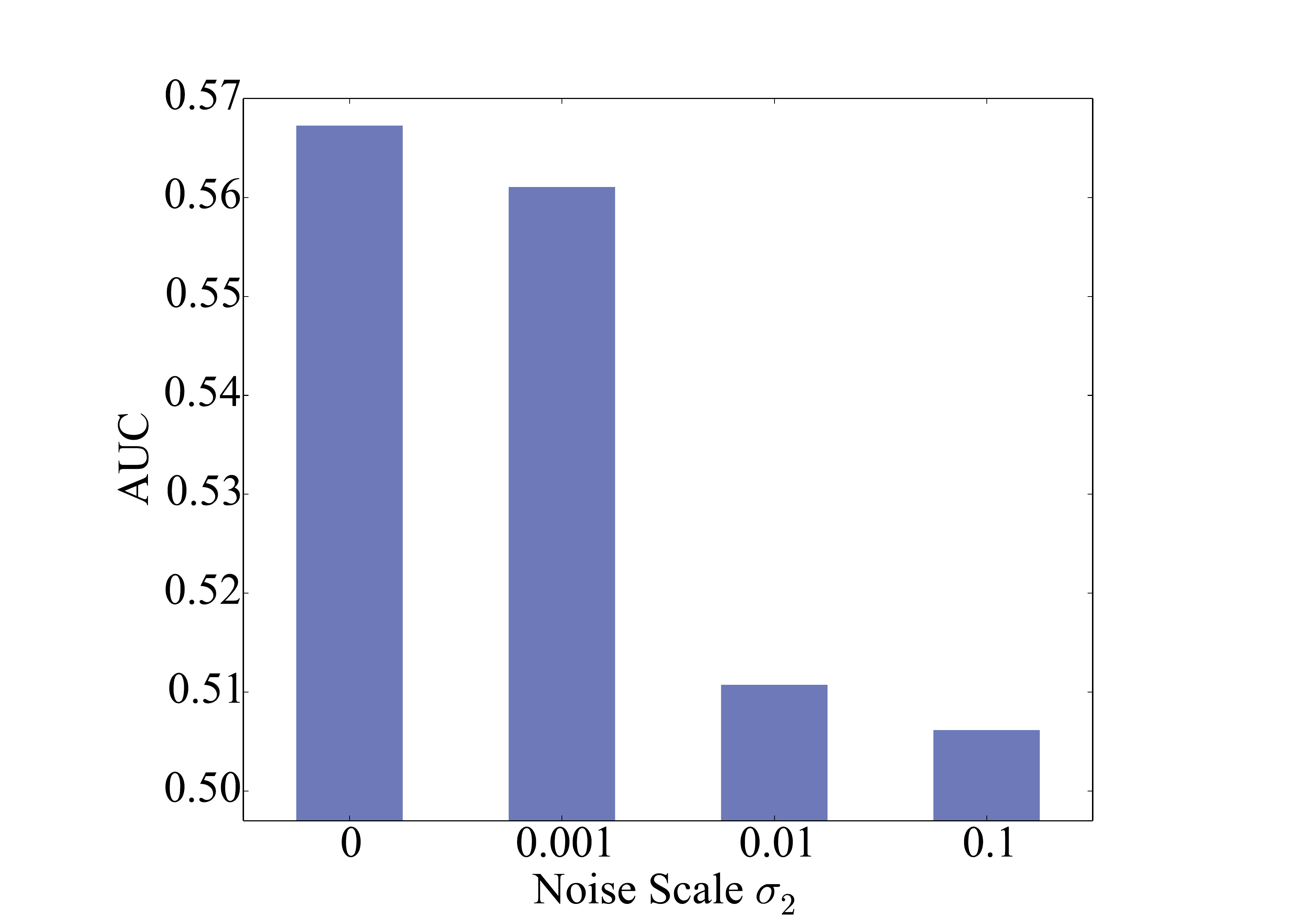}}%
\caption{Performance in terms of area under the curve (AUC) with two types of strategic users: (a) free riders in different percentages, (b) privacy-preserving users in different noise scales $\sigma_2$. } 
\label{fig:randomuser}
\end{figure}

To observe the effects of these strategic behaviors on the model training in FL, we have conducted two experiments on Deep Interest Network (DIN) \cite{zhou2018deep}, a deep learning-based click-through rate (CTR) prediction model, with an industrial dataset from Taobao. We adopt the classic performance metric in machine learning: area under the curve (AUC), and the baseline of AUC is 0.5. The detailed experimental setup can be found in Section \ref{experiment setting}. In Fig. \ref{fig:randomuser}(a), we investigate the effect of free riders, where a free rider randomly generate her model parameter $\hat{x}_p^i \sim \mathcal{N}(0,\sigma_1^2)$, i.e., the Gaussian distribution with variance $\sigma_1=0.01$. As shown in Fig. \ref{fig:randomuser}(a), just 25\% free riders are enough to degrade the performance substantially.
In Fig. \ref{fig:randomuser}(b), we investigate the performance with privacy-preserving behaviors. Each user further strengthen the privacy by adding noises into the model parameters, i.e., $\hat{x}_p^i = x_p^i + \mathcal{N}(0,\sigma_2^2)$, where $\mathcal{N}(0,\sigma_2^2)$ is Gaussian distribution with variance (noise scale) $\sigma_2$. The result in Fig. \ref{fig:randomuser}(b) shows that the AUC depends heavily on the noise scale $\sigma_2$: when $\sigma_2=0.001$, the performance has a slight degradation, but when $\sigma_2=0.01$ or larger, the AUC drops significantly, which demonstrates the harmfulness of overly privacy-preserving behaviors (due to excessive noise). 
With these observations, a contribution evaluation method is highly needed to detect and suppress the free riding behaviors and overly privacy-preserving behaviors. 

To formulate the above strategic behaviors in a unified manner, we define the strategy of a user as $F_{r,a} = P(\hat{x}_p=r|x_p=a)$ for any $r,a \in \{1,2,...,h\}$ and parameter $p$, where $r$ is the reported parameter value and $a$ is the true value.\footnote{Another possibility is to consider the probability conditional on parameter index $p$. However, it is usually not possible for a user to figure out the relation between a particular parameter in a neural network and her own interest, due to the complex structure of neural network models.} 
In other words, the strategy is the probability of reporting $r$ when the value of parameter $p$ is $a$.  
Next, we formulate the above behaviors as informed and uninformed strategies.

\begin{mydef}[\textbf{Informed and Uninformed Strategies}]
A strategy is an uninformed strategy if it has $F_{r,a} = F_{r,b}$ for any $r, a, b \in \{1,2,...,h\}$, otherwise is an informed strategy if there exists $F_{r,a} \neq F_{r,b}$ for some $r, a, b$, it is an informed strategy.
\end{mydef}

Intuitively, the uninformed strategy captures the free-rider behaviors, where the user does not spend resources or effort in training the model, but just report some random parameter values. Hence, the probability of the reported values does not depend on the true values. On the other hand, the informed strategy captures the privacy-preserving behaviors where the user has indeed trained the local model using her local data, but then obfuscated the true parameter value $x^i_{p}$ into a reported value $\hat{x}^i_{p}$ with some probability. 
Note that truthfully reporting model updates is also an informed strategy, where $F_{a,a}=1$  and $F_{r,a}=0, \forall r\ne a$.

\subsection{Peer Prediction and Correlated Agreement}

First introduced in \cite{miller2005eliciting}, the peer-prediction method is a classic mechanism for information elicitation problems without a ground truth, and has been employed in many scenarios, such as crowdsourcing \cite{kamar2012incentives} and peer grading \cite{dasgupta2013crowdsourced}. The key idea of peer prediction is to compare the reported data of user $i$ with that of other users. If their data satisfy some statistical correlation, i.e., the data of user $i$ is predictive of the data of others, then her data is deemed to have a larger contribution, and vice versa. Peer prediction is naturally suitable for FL since there is no ground truth for the contribution evaluation.

For consistency with the terminology used in the peer-prediction literature, we also call each parameter in a FL model, indexed by $p$, a \emph{task}. Each user fulfills a task by training the parameter with her private local data (using, e.g., the gradient descent algorithm). Note that in FL, the task set of each user is the same since all the users are training the same model. We call the value after performing the training task, $x^i_p$, a \emph{signal}, whereby each signal belongs to the set $\{1,2,...,h\}$.

Next, we define a \emph{delta matrix} $\Delta$, which is an $h\times h$ matrix that captures the correlation between each pair of users on a certain task. We first consider the case of homogeneous user, i.e., the delta matrix is the same for each pair of the users. Thus, we omit the user index for the delta matrix. Let $P(a,b)$ denote the joint probability that one user gets signal $a$ after training and the other user gets signal $b$ on the same parameter, and $P(a)$ and $P(b)$ denote the corresponding marginal probabilities. An entry $\Delta(a,b)$ is then defined as
$$\Delta(a,b)\triangleq P(a,b)-P(a)P(b).$$
If $\Delta(a,b)>0$, we have that the signals $a$ and $b$ are positively correlated; if $\Delta(a,b)=0$, we have that $a$ and $b$ are independent; otherwise, they are negatively correlated. In addition, it can be easily verified that the sum of $\Delta(a,b)$ in each row or each column is always 0. We define $Sign(\Delta(a,b)) = 1$ if $\Delta(a,b)>0$, and $Sign(\Delta(a,b)) = 0$ otherwise. Without loss of generality, we assume that there is at least one element in the matrix that is non-zero.

With the above definitions, we describe the \emph{correlated agreement} (CA) method introduced by \cite{dasgupta2013crowdsourced,shnayder2016informed} in the context of FL.
\begin{mydef}[\textbf{CA Method for Homogeneous Users}]
The CA method entails the following steps:
\begin{enumerate}
\item Randomly divide the parameters into a bonus parameter set $M_1$ and a penalty parameter set $M_2$.
\item For each user $i$ and each bonus parameter $p \in M_1$, randomly pick a user $j \ne i$ as the \emph{peer} of $i$, and randomly choose two different penalty parameters $q, q' \in M_2$ for users $i$ and $j$, respectively.
\item \label{step3} The contribution or quality of parameter $p$ of user $i$ is evaluated by $Q^i_p = S(\hat{x}^i_p, \hat{x}^j_p) - S(\hat{x}^i_q, \hat{x}^j_{q'})$, where the score matrix is $S=Sign(\Delta)$, and user $i$'s contribution is $Q^i = \frac{1}{|M_1|} \sum_{p\in M_1}Q^i_p$.
\end{enumerate}
\end{mydef}

The intuition in Step \ref{step3} is that if the reported signals by user $i$ and her peer $j$ on the same parameter $p$ exhibit a positive statistical correlation (e.g., they both report $a$ as in the above example), we give user $i$ a reward of $1$. On the other hand, if their signals on two different parameters $q$ and $q'$ are positively correlated, it suggests that the user $i$ may have randomly uploaded an arbitrary signal for each parameter without actual training, so we give her a penalty of $-1$.

\section{Pairwise Correlated Agreement (PCA)}
\label{PCA}

\begin{algorithm}[t]
\caption{ComputeDelta($i, j, A, B$)}
\label{alg5}
\KwIn{The focus user $i$, the peer user $j$, and the parameter sets $A$ and $B$.}
\KwOut{The delta matrices $\Delta^{i,j}_A, \Delta^{i,j}_B$.}

\For{each pair of signals $a,b \in \{1,2,...,h\}$}{
\label {for}
$T^{i,j}_A(a,b)\leftarrow \frac{\sum_{p\in A} \boldsymbol {1}({\hat{x}^i_p=a, \hat{x}^j_p=b})}{|A|}$.
\label{start_1}

$T^i_A(a)\leftarrow \frac{\sum_{p\in A} \boldsymbol {1}({\hat{x}^i_p=a})}{|A|}$.

$T^j_A(b)\leftarrow \frac{\sum_{p\in A} \boldsymbol {1}({\hat{x}^j_p=b})}{|A|}$.
}
\label {endfor}
Repeat Lines \ref{for} - \ref{endfor} for parameter set $B$.
\label {repeat}

\For{each pair of signals $a,b \in \{1,2,...,h\}$}{
\label {stdel}
$\Delta^{i,j}_A(a,b) \leftarrow T^{i,j}_A(a,b)-T^i_A(a)T^j_A(b)$.

$\Delta^{i,j}_B(a,b) \leftarrow T^{i,j}_B(a,b)-T^i_B(a)T^j_B(b)$.
}
\label {enddel}
\end{algorithm}

The above CA method assumes that all the users are homogeneous, and the delta matrix is the same for all the user pairs. However, one distinctive feature of FL is that users are heterogeneous with non-i.i.d. or imbalanced data \cite{mcmahan2017aistat}. For instance, in mobile recommender systems, the click behaviors of users can be substantially different. 
The work \cite{agarwal2017peer} extended the CA method to the heterogeneous users setting, which divides users into several groups and computes the delta matrix for each pair of groups. However, when there are a large number of heterogeneous users, which is likely to occur in FL, the number of groups would be quite huge, and then this method suffers from a prohibitive computing complexity for both the clustering procedure and the calculation of delta matrices.
To address this issue, we propose a {\em pairwise correlated agreement} (PCA) method, and explores the underlying internal correlation of signal distributions to evaluate the contribution of heterogeneous users.

\begin{algorithm}[t]
\caption{Pairwise Correlated Agreement (PCA) Method for Heterogeneous Users}
\label{alg_PCA}
\KwIn{The set of selected users $K_t$, the reported model updates $\hat{\boldsymbol {x}}^i$ of each user $i \in K_t$, and the number of peers $m$.}
\KwOut{The contribution $Q^i$ of each user $i \in K_t$.}

Randomly divide model parameters into bonus parameter set $M_1$ and penalty parameter set $M_2$.

\For{each user $i \in K_t$}{
$PR^i \leftarrow$ (a set of randomly selected $m$ users in $K_t\backslash i$ as peers).
\label{peer}

\For{each peer $j \in PR^i$}{
Randomly divide parameters into two sets $A, B$ of the same size.
\label{divide1}

$\Delta^{i,j}_A, \Delta^{i,j}_B\leftarrow$ ComputeDelta($i, j, A, B$).
\label{divide2}

\For{each parameter $p \in M_1 \cap A$}{
\label{startA}
Randomly choose two different parameters $q,q'\in M_2\cap A$.

$Q^{i,j}_p \leftarrow Sign(\Delta^{i,j}_B(\hat{x}^i_p, \hat{x}^j_p))-Sign(\Delta^{i,j}_B(\hat{x}^i_q, \hat{x}^j_{q'}))$.
}
\label{endA}

Repeat Lines \ref{startA} - \ref{endA} for each parameter $p\in M_1\cap B$ with $\Delta^{i,j}_A$.
}

$Q^i \leftarrow \frac{1}{m|M_1|} \sum_{p \in M_1}\sum_{j\in PR^i}Q^{i,j}_p$.
\label{calcuQ}
}

\end{algorithm}

Before introducing PCA, we make a reasonable assumption that the signal of different parameters are independent and identically distributed (i.i.d.), which is based on our observation from practical data sets. To validate this assumption, we again conduct two experiments on DIN with an industrial dataset from taobao. We first test the \emph{Spearman's Correlation} \cite{spearman1961proof} between the signals of two randomly chosen parameters to validate the independence hypothesis. Second, we conduct the \emph{Kolmogorov-Smirnov Test} \cite{massey1951kolmogorov} between them to validate the hypothesis of identical distribution. The resulting $p$-values are 0.185 and 0.343, respectively, and both of them are larger than 0.05, which adequately supports the i.i.d. assumption.
We note that this assumption is also used in many related works on sensitivity analysis of neural networks \cite{piche1995selection, soula2006spontaneous, yang2010computing}. 

Under this assumption, we conduct Algorithm \ref{alg5} to estimate the delta matrix for each pair of heterogeneous users. As shown later in the proof of Theorem \ref{thm_error}, the estimation error could be arbitrarily small with a large number of samples.
In algorithm \ref{alg5}, parameters are split into two sets, $A,B$, of the equal size, and we compute a delta matrix for each parameter set and each pair of user $i$ and her peer $j$. The function $ \boldsymbol {1}(\cdot)$ in Lines \ref{start_1} to \ref{endfor} denotes the indicator function. We use $T^{i,j}_A(a,b)$ as the observed frequency of jointly reporting $a,b$ from users $i,j$ for the parameter set $A$, and $T^i_A(a)$ and $T^j_A(b)$ as the corresponding marginal probabilities (Lines \ref{for} - \ref{repeat}). Lines \ref{stdel} - \ref{enddel} compute the delta matrix of each parameter set $A$ and $B$.

Then, PCA is given in Algorithm \ref{alg_PCA}. For each focus user $i$, we randomly choose $m$ peers (Line \ref{peer}). With a larger $m$, the accuracy of $Q^i$ could be improved while the expectation remains the same since we would take the average among them.
Then, for each parameter set $A, B$ of user $i$ and her peer $j$, we apply Algorithm \ref{alg5} to obtain different estimated delta matrices $\Delta^{i,j}_{B}$ and  $\Delta^{i,j}_{A}$, respectively (Lines \ref{divide1} -  \ref{divide2}). Note that we use swapped statistical correlations, i.e., the delta matrices, to calculate the score matrix $Q$ (Line \ref{endA}). As such, the score matrix is independent of the reported parameters themselves.\footnote{To understand this, suppose a signal $r$ occurs only once on parameter $p$. Then since we compute the delta matrix statistically, the signal $r$ will always receive a reward of 1 on this parameter because the correlation is calculated with itself.} 
Lastly, we compute the overall contribution of user $i$'s model update in Line \ref{calcuQ}.

Intuitively, PCA estimates and takes advantage of the correlation between the uploaded models of heterogeneous users.

\section{Applications of PCA}
In this section, we describe two potential applications of PCA: weight calculation in model aggregation and incentive mechanism design. 

\subsection{Weight Calculation for Model Aggregation}

\begin{algorithm}[t]
\caption{Fed-PCA: An improved FL aggregation algorithm using weights calculated by PCA}
\label{alg4}
\KwIn{The set of users $U$, the initialized model $\boldsymbol {M}_0$, the number of learning rounds $T$.}
\KwOut{The global model $\boldsymbol {M}_T$.}
\For{each round $t\in \{1,2,...,T\}$}{
$K_t \leftarrow$ (a set of $k$ users randomly selected from $N$).

\For{each user $i \in K_t$  \textbf{in parallel}}{
Obtain new local model $\boldsymbol {M}_t^i$ and model update $\boldsymbol{x}^i=\boldsymbol {M}_t^i-\boldsymbol {M}_{t-1}$.

Report $\hat{\boldsymbol{x}}^i$ to the server.
}
Get the contribution $Q^i$ for each user $i\in K_t$ by Algorithm \ref{alg_PCA}.

\For{each user $i \in K_t$}{
$w^i \leftarrow \exp(\alpha Q^i) / \sum^k_{j=1} \exp(\alpha Q^j)$.
\label{calcuw}
}
\For{each parameter $p$ in the model}{
\label{agg_start}
$\overline{x}_{p} \leftarrow$ $w^i\hat{x}^i_{p}$.
}
$\boldsymbol {M}_t \leftarrow \boldsymbol {M}_{t-1} + \overline{\boldsymbol {x}}$.
}
\label{agg_end}
\end{algorithm}

In order to reduce the negative effects of low-quality uploaded models, we leverage the contribution value provided by PCA in model aggregation and propose the improved FL algorithm called Fed-PCA.
In particular, Fed-PCA aims to achieve two goals: 1) to prevent falsification of the data size $n^i$, which is a serious issue in FL since it directly affects the weights in model aggregation; 2) to improve the performance of FL as the traditional model aggregation does not take into account the model/data contribution of users. 

As shown in Algorithm \ref{alg4}, we map the value of user $i$'s contribution from the interval $[-1,1]$ to a non-negative weight $w^i = \exp(\alpha Q^i) / \sum^k_{j=1} \exp(\alpha Q^j)$ in Line \ref{calcuw}. Here, we use the exponential function with a controlled parameter $\alpha$, which determines the variance of user weights in model aggregation. 
The local models are then aggregated into the global model using the new normalized user weights (Lines \ref{agg_start} - \ref{agg_end}). Note that our contribution values could be used in not only the weighted averaging aggregation process as shown here, but also the recent median-based mechanisms proposed to defend Byzantine attacks \cite{yin2018byzantine,chen2019distributed}. We evaluate the performance for both the averaging and median-based aggregation methods in Section \ref{Performance Evaluation}.

\subsection{Incentive Mechanism Design}

We next discuss how to prevent the strategic behaviors through incentives, where an essential property for this issue is \emph{incentive compatibility}.
We say that incentive compatibility is satisfied when it is in all users' best interests (obtaining the highest rewards) to report their true models, i.e.,
$$U^i(\boldsymbol {x}^i, \{\hat{\boldsymbol {x}}^j\}^{j\neq i})\ge U^i(\hat{\boldsymbol {x}}^i, \{\hat{\boldsymbol {x}}^j\}^{j\neq i})$$ 
for any reported $\hat{\boldsymbol {x}}^i$, where $U^i$ denotes the utility obtained by user $i$, which will be defined later, and $\{\hat{\boldsymbol {x}}^j\}^{j\neq i}$ is the set of reported model updates of all the other users. We emphasize that incentive compatibility could be quite important for FL, because otherwise, users may upload falsified models and the learning problem would be ill-defined. 

Next, based on the definition of informed and uninformed strategies in Section \ref{strategy}, we introduce the concept of informed incentive compatibility and $\epsilon$-informed incentive compatibility for the above defined strategic behaviors. We denote by $\mathbb{I}^i$ and $\{\mathbb{I}\}^{j\neq i}$ the truthful strategy of user $i$ and that of other users, respectively.

\begin{mydef}[\textbf{Informed Incentive Compatibility and $\epsilon$-Informed Incentive Compatibility}]
If for every user $i$ and any informed strategies $F, G$, and some $\epsilon\ge0$, we have $$U^i(\mathbb{I}^i,\{\mathbb{I}\}^{j\ne i}) \ge U^i(F^i,\{G\}^{j\ne i})-\epsilon,$$ and for any uninformed strategy $F'$, we have $$U^i(\mathbb{I}^i,\{\mathbb{I}\}^{j\ne i}) > U^i(F'^i,\{G\}^{j\ne i}),$$ 
then the mechanism is $\epsilon$-informed incentive compatible. If $\epsilon=0$, the mechanism is informed incentive compatible.
\end{mydef}

The definition of $\epsilon$-informed incentive compatibility reflects that (i) if a user adopts an uninformed strategy, her contribution is strictly lower than that of truthful reporting; (ii) if a user adopts an informed (yet still untruthful) strategy, her contribution is at best $\epsilon$ more than that of truthful behavior, where $\epsilon$ is a small constant number. Therefore, an incentive mechanism with this property can prevent strategic users from uninformed or informed strategies.

Aiming to illustrate the application of PCA in a clearest manner, we assume that the utility of each user is simply her received reward, regardless of other factors, that is, $U^i(\hat{\boldsymbol {x}}^i, \{\hat{\boldsymbol {x}}^j\}^{j\neq i}) = R^i(\hat{\boldsymbol {x}}^i, \{\hat{\boldsymbol {x}}^j\}^{j\neq i})$ where $R^i(\hat{\boldsymbol {x}}^i, \{\hat{\boldsymbol {x}}^j\}^{j\neq i})$ is the reward received by user $i$.
It should be noted that PCA could be easily integrated into most of other existing incentive mechanisms, such as that in \cite{yu2020fairness,zeng2020fmore}, and more factors could be taken into account, e.g., the training cost, the privacy cost and the budget balance property.

On top of this simplified utility model of users, we propose that the following simple mechanism could guarantee the $\epsilon$-informed incentive compatibility: allocating a reward proportional to her contribution, i.e., $R^i(\hat{\boldsymbol {x}}^i, \{\hat{\boldsymbol {x}}^j\}^{j\neq i})= f( Q_i(\hat{\boldsymbol {x}}^i, \{\hat{\boldsymbol {x}}^j\}^{j\neq i}))$ where $f$ could be any positive monotonic increasing function.
We posit the following main theorem, which shows that this mechanism is close to informed incentive compatibility.
\begin{thm}
\label{thm_error}
Let $\epsilon > 0$, and $\delta > 0$. If the number of samples is $g=O(9h^2\log(1/\delta)/\epsilon^2)$, then with a probability of at least $1-\delta$, the above incentive mechanism with heterogeneous users is $\epsilon$-informed incentive compatible.
\end{thm}

We omit the proof due to space constraint.

\section{Performance Evaluation}
\label{Performance Evaluation}
In this section, we report the evaluation results of our proposed approach and methods (PCA, Fed-PCA, and incentive mechanism) using the MNIST dataset and an industrial product recommendation dataset. 

\begin{figure}[t]
\centering
\subcaptionbox{}{\includegraphics[width=0.23\textwidth,trim=100 0 150 60,clip]{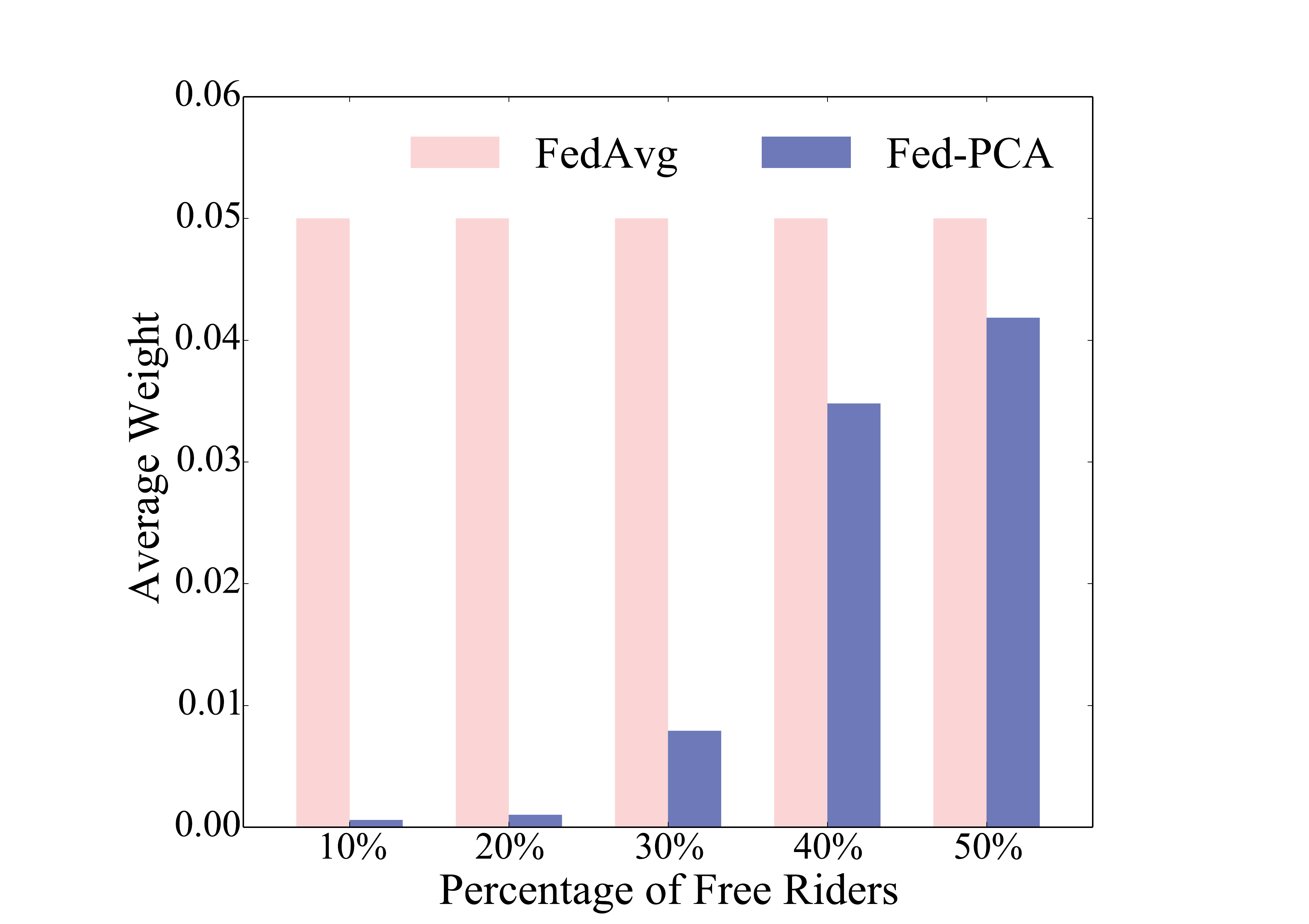}}%
\hfill
\subcaptionbox{}{\includegraphics[width=0.23\textwidth,trim=100 0 150 60,clip]{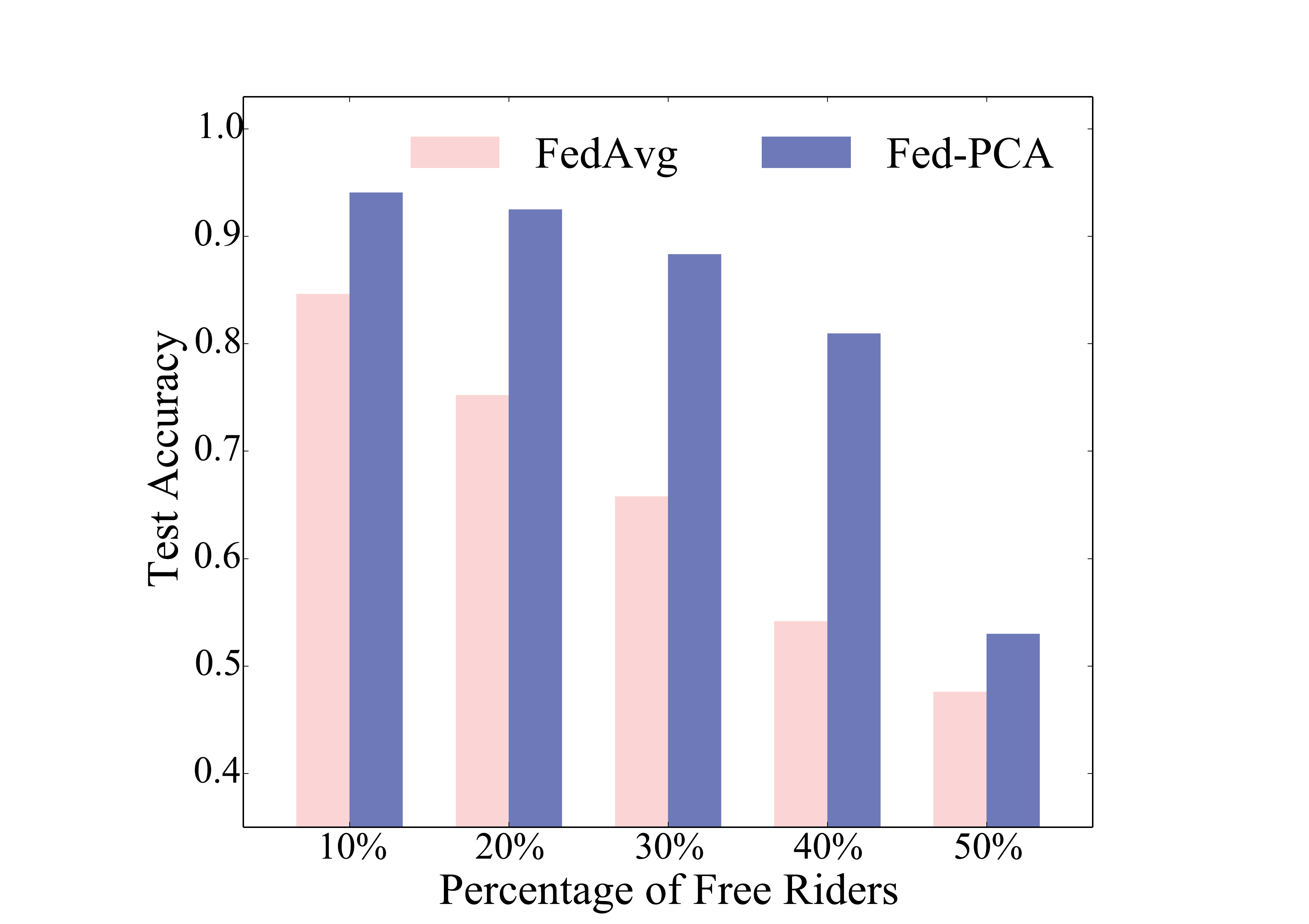}}%
\caption{Results on MNIST with free-riding users: (a) Average weight assigned to free-riding users; (b) Test accuracy.}
\label{fig:mnist_freerider}
\end{figure}

\subsection{Experiments on the MNIST Dataset}
\textbf{Experimental Setting.}
We first study the performance of Fed-PCA for the MNIST digit-recognition dataset using a multi-layer perceptron with a hidden layer of 100 units. ReLU activations and dropout technique are adopted in the experiments. The network contains a total of 159,010 parameters. 
Similar to \cite{mcmahan2017aistat}, we first sort the MNIST data by digit label, and divide them into 200 shards, each of which includes 300 data items. We assume there are 100 users in total in the FL system, each of which is randomly assigned two shards of data. This way, the data distribution among users is highly non-i.i.d.\footnote{Following \cite{mcmahan2017aistat}, we adopt this balanced partition, but in our experiments on the industrial dataset, users are divided naturally in an unbalanced manner.} 

For the model updates, we choose a gradient quantization technique called cpSGD \cite{agarwal2018cpsgd} to reduce the communication cost. 
In the FL training, $k=20$ users are randomly chosen in one round. We adopt mini-batch stochastic gradient descent (SGD) with momentum as the optimization algorithm. The value of momentum is set to 0.5, the batch size is 10, the local epoch number in each round is 5, and the learning rate is 0.01. In the quantization process, we set $h=8$, and $X^{max}=0.1$. In PCA, the peer number $m$ is 5, and the number of bonus parameters is $|M_1| = 1,000$. The parameter $\alpha$, which converts the user contribution into her weight, is set to 10. Regarding the free riders in the system, we assume the parameters they generate in their model updates to be $\hat{x}_p^i=\mathcal{N}(0,\sigma_1^2)$, where $\sigma_1$ is set to 0.01. In the experiments on  privacy-preserving users, we assume that 25\% of users add noises into their model parameters, each of which reports $\hat{x}_p^i = x_p^i +  \mathcal{N}(0,\sigma_2^2)$. We will test the impact of different noise scales $\sigma_2$.

\begin{figure}[t]
\centering
\subcaptionbox{}{\includegraphics[width=0.23\textwidth,trim=100 0 150 60,clip]{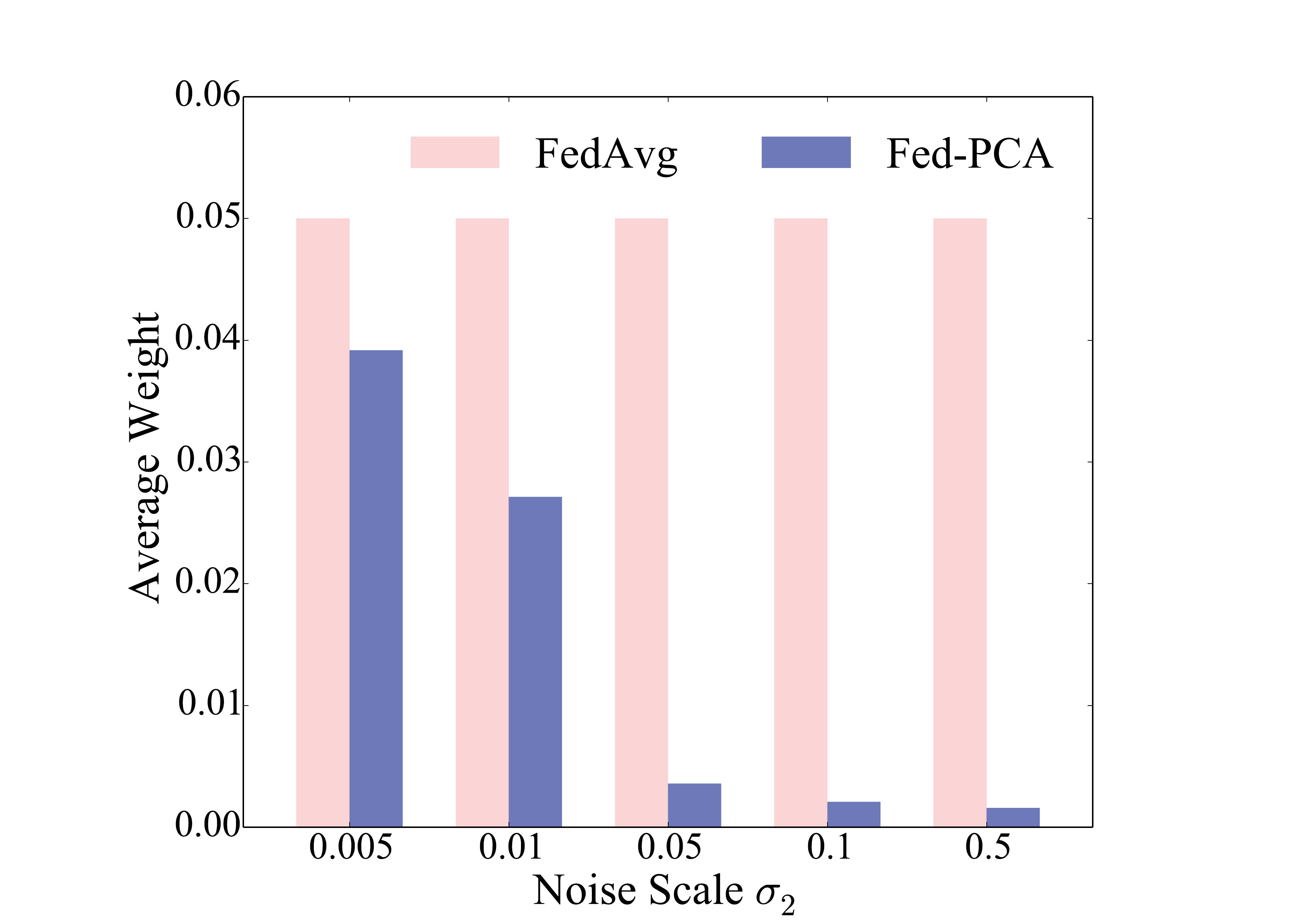}}%
\hfill
\subcaptionbox{}{\includegraphics[width=0.23\textwidth,trim=100 0 150 60,clip]{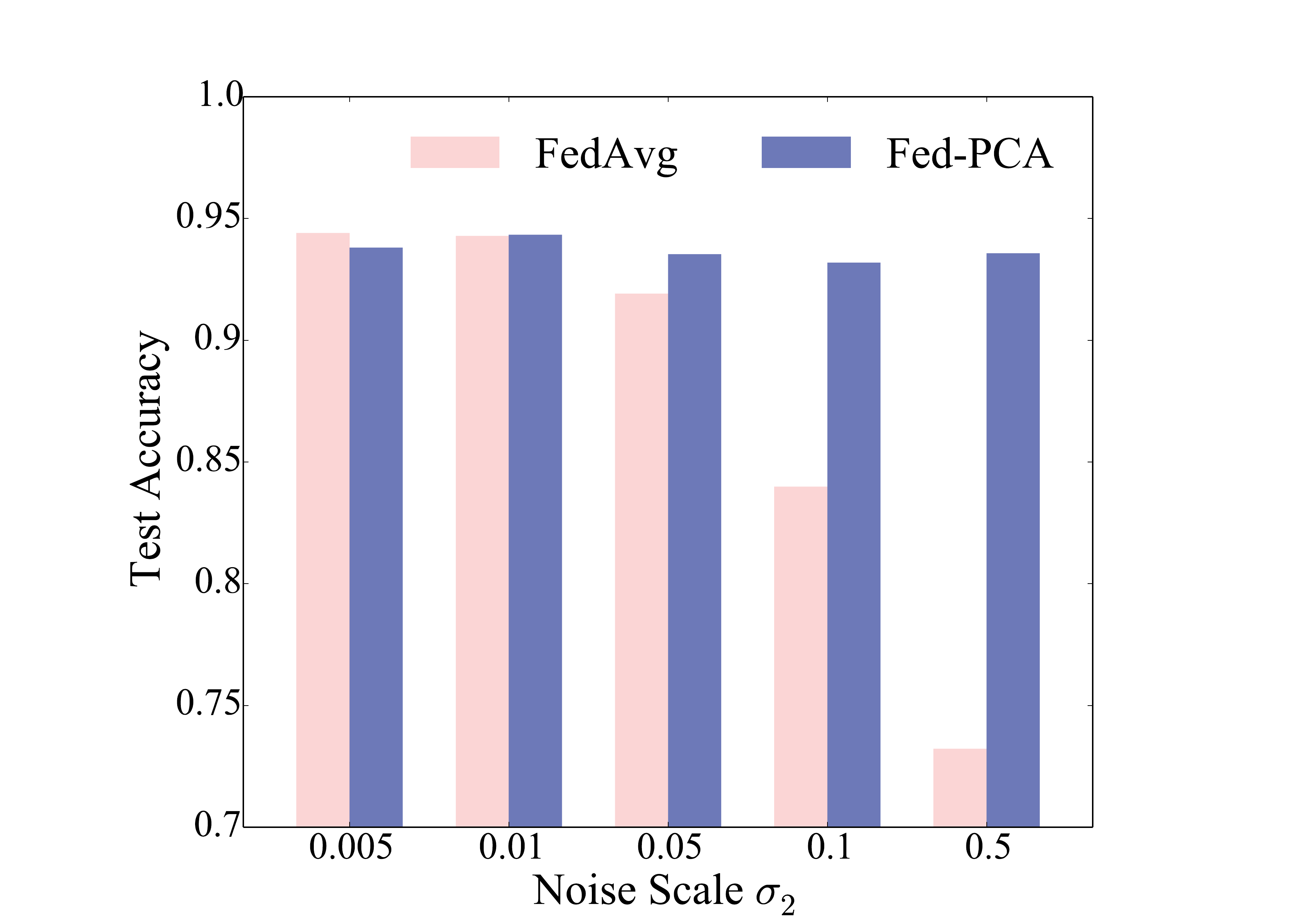}}%
\caption{Results on MNIST with (overly) privacy-preserving users: (a) Average weight assigned to privacy-preserving users. (b) Test accuracy with 25\% privacy-preserving users.}
\label{fig:mnist_noise}
\end{figure}

\textbf{Experimental Results.}
We first compare the performance of our Fed-PCA with the original FedAvg algorithm with different percentages of free riders. For clarity, we illustrate the user contributions using their corresponding weights, as calculated in Algorithm \ref{alg4}. Fig. \ref{fig:mnist_freerider}(a) shows the average weight of free riders. When the percentage of free riders is no more than 20\%, PCA evaluates their contributions (and hence weights) as nearly 0, while FedAvg constantly assigns them a weight of 0.05 (recall there are 20 users). This result demonstrates that free riders are detected accurately by PCA, while further implying the effectiveness of our incentive mechanism. When the percentage of free riders increases to 50\%, their average weight increases accordingly, but is always lower than that of FedAvg. This is because that the model quality of peers decreases with the percentage of free riders.
Note that we could assume most users in FL are likely to be truthful, which provides a reference to detect free riders. This is a mild assumption in real life and has been widely adopted in previous literature \cite{lin2019free,munoz2019byzantine}.
In Fig. \ref{fig:mnist_freerider}(b), we study the test accuracy with different percentages of free riders. It is depicted in the figure that Fed-PCA decisively outperforms the FedAvg algorithm. The reason is that PCA detects free riders successfully and, thus, assigns them low weights in the model aggregation process. We also note that in the normal case without any free riders or privacy-preserving users, the accuracy of Fed-PCA is 0.9421 after 100 rounds, which is quite close to the baseline of FedAvg: 0.9458.

The impact of the noise scale $\sigma_2$ of privacy-preserving users is shown in Fig. \ref{fig:mnist_noise}. Fig. \ref{fig:mnist_noise}(a) explores the average weight of privacy-preserving users. When the noise scale $\sigma_2$ is no less than 0.05, the average weight of privacy-preserving users is nearly 0, while a lower noise scale leads to a higher weight. This result demonstrates that PCA is capable of detecting users who add large noises, but ignores the slight noises that do not affect the performance of FL, which is a good characteristic of the corresponding incentive mechanism. In Fig. \ref{fig:mnist_noise}(b), we show the test accuracy of Fed-PCA and FedAvg. The accuracy of FedAvg degrades seriously with the increase of $\sigma_2$, but our Fed-PCA remains nearly unaffected by users with large noises because they are assigned very low weights.

\subsection{Experiments on the Industrial Dataset}

\begin{figure}[t]
\centering
\subcaptionbox{}{\includegraphics[width=0.24\textwidth,trim=100 0 150 60,clip]{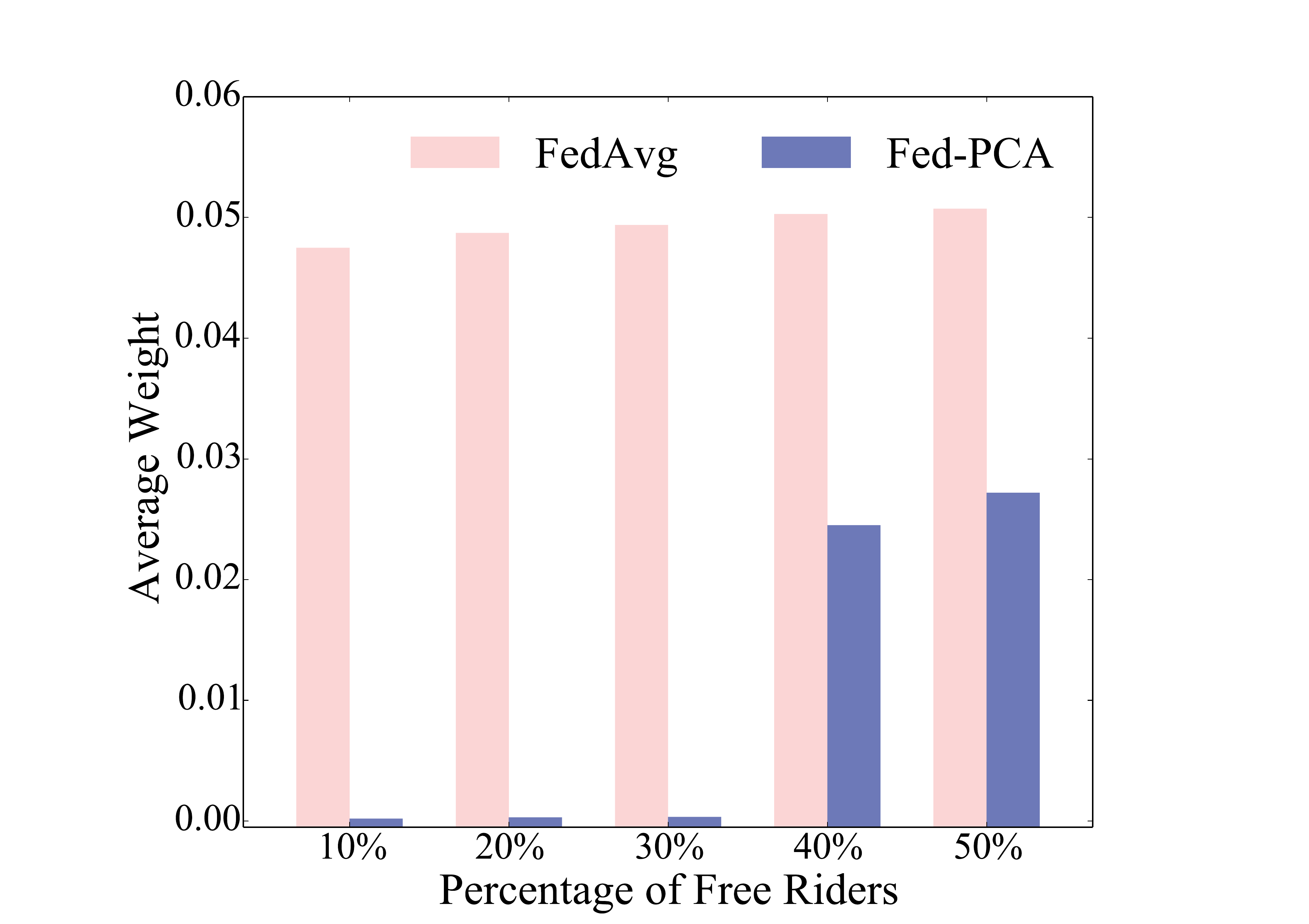}}%
\hfill
\subcaptionbox{}{\includegraphics[width=0.24\textwidth,trim=100 0 150 60,clip]{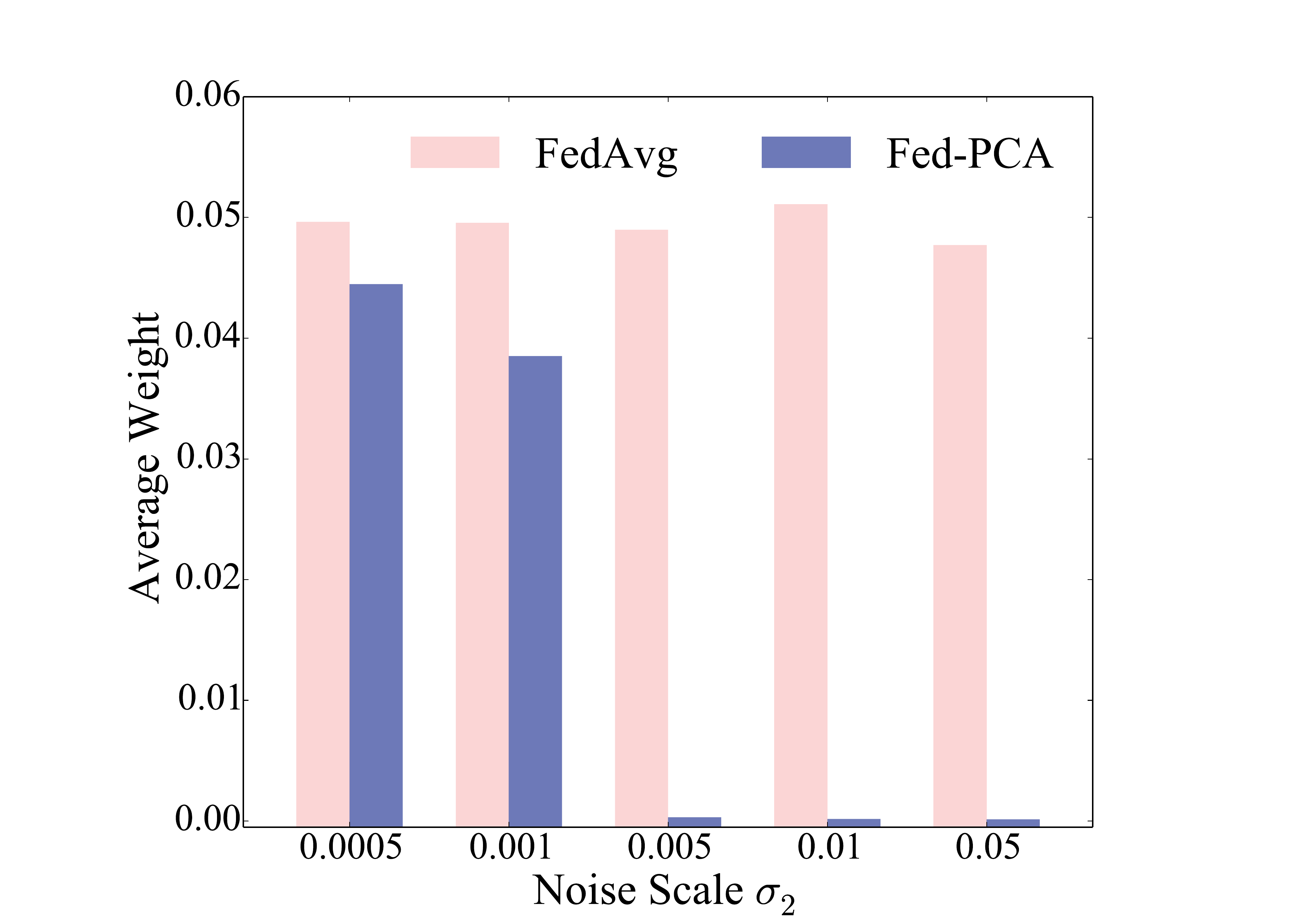}}%
\caption{Results on industrial dataset: (a) Average weight assigned to free-riding users. (b) Average weight assigned to privacy-preserving users.}
\label{fig:taobao_weight}
\end{figure}

\textbf{Experimental Setting.}
\label{experiment setting}
We next evaluate the performance of our Fed-PCA on an industrial dataset from Taobao, one of the largest a mobile recommendation system of products in China. In the dataset, there are 30-day impressions and click logs of users (dated from June 15 to July 15, 2019).
We use the Deep Interest Network (DIN) \cite{zhou2018deep} as the machine learning models. We refer the readers to \cite{zhou2018deep} for more details on DIN and the dataset.

In the training of FL, the batch size is set to 2, the epoch number is set to 1, and the learning rate is initialized as 1.0 with an exponential decay rate of 0.999. In the quantization process, we set $h=256$, and set $X^{max}$ as 0.1 multiplied by the learning rate in the round. In PCA, the parameter $\alpha$ is set to 100 if not otherwise stated. 
The other parameters are the same as those of the experiments on the MNIST dataset.

\begin{figure}[t]
\centering
\includegraphics[width=0.32\textwidth,trim=0 0 0 80,clip]{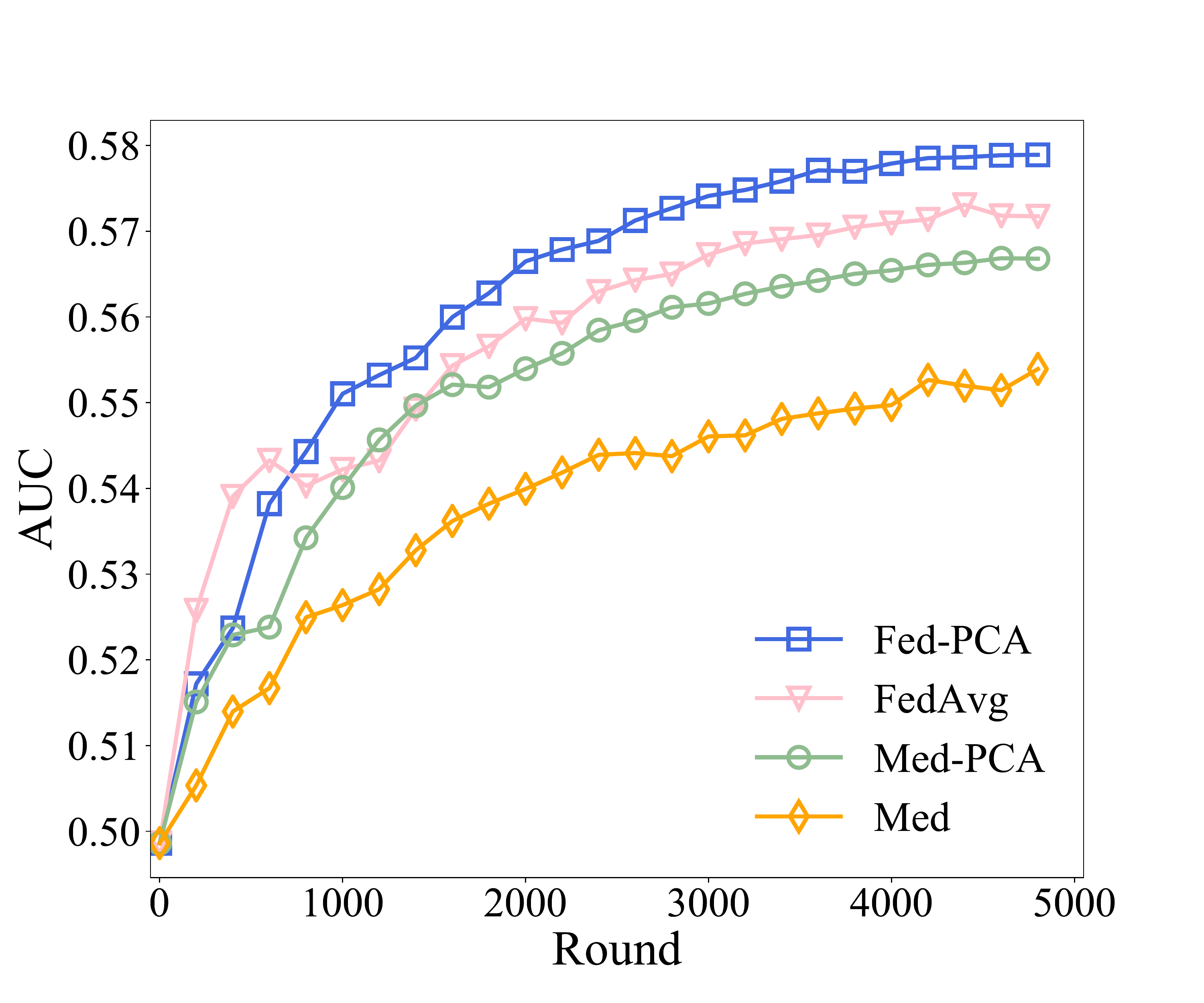}
\caption{Comparison of different model aggregation methods.}
\label{fig:method}
\end{figure}

\textbf{Experimental Results.}
We first explore the average weight of free riders and privacy-preserving users in Fig. \ref{fig:taobao_weight}. 
In Fig. \ref{fig:taobao_weight}(a), we can see that when the percentage of free riders is no more than 30\%, all of them are detected in PCA and, hence, are allocated merely 0 weight in the aggregation. When the percentage increases to 50\%, Fed-PCA still has the ability to detect them to some extent; thus, their average weight is about half that of truthful users. We investigate the impact of privacy-preserving users in Fig. \ref{fig:taobao_weight}(b). A clear noise scale boundary of 0.001 is found, above which the user would be detected and punished with a low weight. These results clearly suggest the effectiveness of PCA in the application of incentive mechanism design.

Then, we test the AUC with different aggregation methods in the system consisting of exclusively truthful users in Fig. \ref{fig:method}. As median-based aggregation methods has attracted much attention in recent years as resistant to the attacks of malicious users \cite{yin2018byzantine,munoz2019byzantine}, we conduct experiments on both averaging and median aggregation.
In the figure, the black line (Med) is the performance of the unweighted median-based aggregation method, and the red line (Med-PCA) represents a mix of PCA and the weighted median methods. 
It is depicted that the unweighted median-based aggregation method underperforms all other approaches since more information of the model updates from the users are lost due to the median operation, compared with the averaging operation. PCA helps alleviate this problem by allocating larger weights to the users with better performance in the median operation.
Surprisingly, we can observe that, under the premise of preventing free riders and overly privacy-preserving users, the performance of Fed-PCA clearly surpasses all other aggregation methods, including FedAvg. This means that the contributions and, hence, the weights calculated by PCA are even more reasonable and effective than the traditional weights directly calculated by data sizes, in absence of the information about local datasets. We conjecture that this is because of the difference between the synthetic MNIST dataset and the industrial dataset. In MNIST, each data item has approximately the same value for the learning problem, and hence the data sizes could be used as a good proxy of weights in model aggregation. But in the industrial dataset, real users may provide many useless data items for the learning problem, such as unintended activations. Therefore, a well-designed weight assignment approach has the potential to evaluate the true contribution of each user, and hence to outperform the traditional FL algorithm.

\section{Related Work}
\label{related}
The FL framework was first introduced by Google \cite{mcmahan2017aistat,konevcny2016federated} as a new paradigm of distributed machine learning, and it has been applied in a virtual keyboard named Gboard \cite{yang2018applied,hard2018federated}. 
Some recent works have taken incentive mechanisms of FL into account \cite{ding2020incentive, yu2020fairness, zeng2020fmore, liu2020fedcoin, jia2019towards, huang2020exploratory}.
For example, \cite{kang2019incentive} proposes an incentive mechanism for FL using the contract theory, by considering the computation and communication costs of training the model.
These studies focused on the costs of users, while the contribution of each user to the FL platform is simplified as a sandbox or a linear function of her cost, which is not so practical in real life. Some other work \cite{wang2019measure, jia2019towards, liu2020fedcoin}  
account for the contribution using the Shapley value-based techniques. However, as stated above, the computation of Shapley values suffers from high time complexity and the requirement of a representative test dataset in FL, which poses obstacles for fair contribution evaluation. Closely related works are \cite{liu2020incentives, weng2020fedserving}, which adopt peer prediction to compare the output of the updated models of users on the test data, and thus a representative test dataset is still required in their proposed approaches.
Our work tackles this problem by leveraging the technique of peer prediction on the model parameters, whereby the contribution of each user is evaluated without either training data or test data.

\section{Conclusion}
\label{conclusion}
In this paper, we have proposed a pairwise correlated agreement (PCA) method to evaluate the contributions of users in federated learning, without requiring a test dataset. 
We have then applied PCA in (1) weight calculation for more robust model aggregation, where we have designed Fed-PCA as a better alternative to FedAvg, and (2) incentive mechanism design for FL.
Extensive experiments are conducted using the MNIST dataset and a large industrial product recommendation dataset. The evaluation results demonstrate the effectiveness of our proposed approach in terms of both detecting strategic user behaviors and improving prediction accuracy. 
\bibliographystyle{IEEEtran}
\bibliography{wiopt}

\end{document}